# SparsityBoost: A New Scoring Function for Learning Bayesian Network Structure

**Eliot Brenner,*  David Sontag**
Courant Institute of Mathematical Sciences
New York University

## Abstract

We give a new consistent scoring function for structure learning of Bayesian networks. In contrast to traditional approaches to score-based structure learning, such as BDeu or MDL, the complexity penalty that we propose is data-dependent and is given by the probability that a conditional independence test correctly shows that an edge cannot exist. What really distinguishes this new scoring function from earlier work is that it has the property of becoming computationally *easier* to maximize as the amount of data increases. We prove a polynomial sample complexity result, showing that maximizing this score is guaranteed to correctly learn a structure with no false edges and a distribution close to the generating distribution, whenever there exists a Bayesian network which is a perfect map for the data generating distribution. Although the new score can be used with any search algorithm, we give empirical results showing that it is particularly effective when used together with a linear programming relaxation approach to Bayesian network structure learning.

## 1 Introduction

We consider a fundamental problem in statistics and machine learning: how can one automatically extract structure from data? Mathematically this problem can be formalized as that of learning the structure of a Bayesian network with discrete variables. Bayesian networks refer to a compact factorization of a multivariate probability distribution, one-to-one with an acyclic graph structure, in which the conditional probability distribution of each random variable depends only on the values of its parent variables. One application of Bayesian network structure learning is for the discovery of protein regulatory networks from gene expression or flow cytometry data (Sachs *et al.* , 2005).

Existing approaches to structure learning follow two basic methodologies: they either search over structures that maximize the likelihood of the observed data (score-based methods), or they test for conditional independencies and use these to constrain the space of possible structures. The former approach leads to extremely difficult combinatorial optimization problems, as the space of all possible Bayesian networks is exponentially large, and no efficient algorithms are known for maximizing the scores. The latter approach gives fast algorithms but often leads to poor structure recovery because the outcomes of the independence tests can be inconsistent, due to sample size problems and violations of assumptions.

We formulate a new objective function for structure learning from complete data which obtains the best of both worlds: it is a score-based method, based predominantly on the likelihood, but it also makes use of conditional independence information. In particular, the new objective has a "sparsity boost" corresponding to the log-probability that a conditional independence test correctly shows that an edge cannot exist. We show empirically that this new objective substantially outperforms the previous state-of-the-art methods for structure learning. In particular, on synthetic distributions we find that it learns the true network structure with less than half the data and one tenth the computation.

The contributions of this paper are the introduction of this new scoring function, a proof of its consistency (we show polynomial sample complexity), and a carefully designed importance sampling algorithm for efficiently computing the confidence scores used in the objective. For both the proof of sample complexity and

*Current Affiliation: Search & Algorithms, Shutterstock Inc., New York

the importance sampling algorithm, we develop several new results in information theory, constructing precise mappings between a parametrization of distributions on two variables and mutual information, and characterizing the rate of convergence of various quantities relating to mutual information. We expect that many of the techniques that we developed will be broadly useful beyond Bayesian network structure learning.

## 2 Background

This paper considers the problem of learning Bayesian network structure from complete data (no hidden variables or unobserved factors). Let $\mathcal{X} = (X_1, X_2, \ldots, X_n)$ be a collection of random variables. For reasons that we explain in the next section, our results are restricted to the case when the variables $X_i$ are binary, i.e. $\mathrm{Val}(X_i) = \{0, 1\}$. Formally, a Bayesian network over $\mathcal{X}$ is specified by a pair $(G, P)$, where $G = (V, E)$ is a directed acyclic graph (DAG) satisfying the following conditions: the nodes $V$ correspond to the variables $X_i \in \mathcal{X}$ and $E$ is such that every variable is conditionally independent of its non-descendants given its parents. The joint distribution can then be shown to factorize as $P(x_1, \ldots, x_n) = \prod_{i \in V} P(X_i = x_i \mid X_{\mathrm{Pa}(i)} = x_{\mathrm{Pa}(i)})$, where $\mathrm{Pa}(i)$ denotes the parent set of variable $X_i$ in the DAG $G$, and $x_{\mathrm{Pa}(i)}$ refers to an assignment to the parents.

A Bayesian network $G$ is called an independence map (I-map) for a distribution $P$ if all the (conditional) independence relationships implied by $G$ are present in $P$. Going one step further, $G$ is called a perfect map for $P$ if it is an independence map and the conditional independence relationships implied by $G$ are the *only ones* present in $P$. By $\omega_N$ (or in some contexts, $Y_N$) we denote a sequence of observations of the random variables $\mathcal{X}$, generated i.i.d. from an unknown Bayesian network $(G, P)$, where $G$ is a perfect map for $P$. The problem that we study is that of learning the Bayesian network structure and distribution $(G, P)$ from the samples $\omega_N$.

The simplest case of learning BN structure is when we have two random variables, which we will call $X_A$ and $X_B$. There are only two nonequivalent BN structures:

$$G_0\colon X_A \qquad X_B \text{ (``disconnected''),}$$
$$G_1\colon X_A \longrightarrow X_B \text{ (``connected'').}$$

The structure learning problem in this case is to return, based on $\omega_N$, a decision $X_A \perp\!\!\!\perp X_B$ $(G_0)$ or $X_A \not\perp\!\!\!\perp X_B$ $(G_1)$. In other words, in this case, the structure learning problem is strictly equivalent to one case of *hypothesis testing*, a well-studied and classic problem in statistics, specifically testing the hypothesis of whether $X_A$ and $X_B$ are independent.

In the case of three or more variables, the equivalence no longer holds in any strict sense. *Constraint-based approaches* use the results of conditional independence tests to infer the model structure. These methods solve the structure learning problem sequentially by first learning the undirected skeleton of the graph, $\mathrm{Skel}(G)$, and then orienting the edges to obtain a DAG. Assuming that $G$ is a perfect map for $P$, if $A$ is conditionally independent of $B$ then we can conclude that neither $A \to B$ nor $B \to A$ can be in $G$. It can be shown that either $A$'s parents or $B$'s parents will be a separating set proving their conditional independence (there may be others). Thus, if we make the key assumption that each variable has at most a fixed number of parents $d$, then this can yield a polynomial time algorithm for structure learning (Spirtes *et al.* , 2001; Pearl & Verma, 1991). However, this approach has a number of drawbacks: difficulty setting thresholds, propagation of errors, and inconsistencies.

Let $p = p(\omega_N, A, B \mid s)$ denote the empirical distribution of $A$ and $B$ conditioned on an assignment $S = s$ for $S \subseteq V \backslash \{A, B\}$, and marginalized over all of the other variables. The *mutual information* statistic,

$$MI(p) = \sum_{a \in \mathrm{Val}(A),\, b \in \mathrm{Val}(B)} p(a, b|s) \log\left(\frac{p(a, b|s)}{p(a|s)p(b|s)}\right)$$

is a measure of the conditional independence of $A$ and $B$ conditioned on $S = s$. Given infinite data, two variables are independent if and only if their mutual information is zero. However, with *finite* data, mutual information is *biased* away from zero (Paninski, 2003). As a result, it can be very difficult to distinguish between independence and dependence.

An alternative approach is to construct a scoring function which assigns a value to every possible structure, and then to find the structure which maximizes the score (Lam & Bacchus, 1994; Heckerman *et al.* , 1995). Perhaps the most popular score is the BIC (Bayesian Information Criterion) score:

$$S_{\psi_1}(\omega_N, G) = LL(\omega_N|G) - \psi_1(N) \cdot |G|. \qquad (1)$$

Here, $LL(\omega_N|G)$ is the log-likelihood of the data given $G$, $|G|$ is the number of parameters of $G$, and $\psi_1(N)$ is a weighting function with the property that $\psi_1(N) \to \infty$ and $\psi_1(N)/N \to 0$ as $N \to \infty$. When $\psi_1(N) := \frac{\log N}{2}$, the score, now called MDL, can be theoretically justified in terms of Bayesian probability. Intuitively, we can explain the BIC/MDL score as a log-likelihood regularized by a *complexity penalty* to keep fully connected models (with the most parameters) from always winning. Finding the structure which maximizes the score is known to be NP-hard (Chickering, 1996; Chickering *et al.* , 2004; Dasgupta, 1999). Heuristic algorithms have been proposed for maximizing this score, such as greedy hill-climbing (Chickering, 2002; Friedman *et al.* , 1999) and, more recently, by formulating the structure learning problem as an integer linear program and solving using branch-and-cut (Cussens, 2011; Jaakkola *et al.* , 2010).

The running time of solving the integer linear programs dramatically increases as the amount of data used for learning increases (see, e.g., Fig 4). This is counter-intuitive: more data should make the learning problem easier, not harder. The core problem is that as the amount of data increases, the likelihood term grows in magnitude whereas the complexity penalty shrinks. This is necessary to prove that these scoring functions are consistent, i.e. that in the limit of infinite data the structure which maximizes the score in fact is the true structure. As a consequence, however, the score becomes very flat near the optimum with a large number of local maxima, making the optimization problem extremely difficult to solve.

## 3 SparsityBoost: A New Score for Structure Learning

We design a new scoring function for structure learning that is both consistent and easy to solve regardless of the amount of data that is available for learning. The key property that we want our new scoring function to have is that as the amount of data increases, optimization becomes easier, not harder. When little data is available, it should reduce to the existing scoring functions.

Our approach is to add, to the BIC score, new terms derived from statistical independence tests. Before introducing the new score we provide some background on hypothesis testing. Let $\mathcal{P}$ denote the simplex of (joint) probability distributions over a pair of random variables, and let $\mathcal{P}_0$ denote the subset of product distributions: $\mathcal{P}_0 = \{q \in \mathcal{P} \mid MI(q) = 0\}$. For $q \notin \mathcal{P}_0$, the magnitude of $MI(q)$ provides a measure of how far $q$ is from the set of product distributions. For $\eta > 0$, we define $\mathcal{P}_\eta := \{q \mid MI(q) \geq \eta\}$. The testing procedure has $\omega_N$ as input, null hypothesis $H_0$ (independence) for $p \in \mathcal{P}_0$, and alternative hypothesis $H_1$ for $p \in \mathcal{P}_\eta$. The *Type I error* $\alpha_N$ is defined as the probability of the test rejecting a true $H_0$, the *Type II error* $\beta_N$ is defined as the probability of the test falsely accepting $H_0$, and the *power* is defined as $1 - \beta_N$.

The theory of Neyman-Pearson hypothesis testing for composite hypotheses tells us how to construct a hypothesis test of maximal *power* for any $\alpha_N$ (Hoeffding, 1965; Dembo & Zeitouni, 2009). In our setting, the test corresponds to computing $MI(\omega_N) := MI(p(\omega_N))$ and deciding on $H_1$ if the test statistic exceeds a threshold $\gamma$. Let $\beta_N(\gamma)$ denote the Type II error of the Neyman-Pearson test with threshold $\gamma$.

We propose using in our score the Type II error of the test with threshold $MI(\omega_N)$,

$$\beta_N^{p^\eta}(MI(\omega_N)) := \Pr_{Y_N \sim p^\eta}\{MI(Y_N) \leq MI(\omega_N)\},$$

where $p^\eta$ is the $M$-projection of $p(\omega_N)$ onto $\mathcal{P}_\eta$, that is, with $H(\cdot\|\cdot)$ denoting the Kullback-Leibler divergence,

$$p^\eta := \operatorname*{argmin}_{p \in \mathcal{P}_\eta} H(p(\omega_N)\|p). \tag{2}$$

An intuitive explanation for the Type II error is that $\beta_N^{p^\eta}(\gamma)$ is the probability of obtaining a test statistic $MI(Y_N)$, $Y_N \sim p^\eta$, that is *more extreme*, in the *wrong* direction of independence, than the observed test statistic $\gamma$. On the one hand, if $\omega_N \sim p_0 \in \mathcal{P}_0$, then with high probability the power of the test with threshold $MI(\omega_N)$ approaches 1 and $\beta_N^{p^\eta}(MI(\omega_N))$ approaches 0, exponentially fast as $N \to \infty$; on the other hand, if $\omega_N \sim p_1 \in \mathcal{P}_\epsilon$, where $\epsilon > \eta$, then with high probability the power approaches 0 and $\beta_N^{p^\eta}(MI(\omega_N))$ approaches 1, as $N \to \infty$.

Now we can state our new score for structure learning and explain its remaining features:

$$S_{\eta,\psi_1,\psi_2}(\omega_N, G) = LL(\omega_N|G) - \psi_1(N)\cdot|G| + \psi_2(N)\cdot \sum_{(A,B)\notin G} \max_{S\in S_{A,B}(G)} \min_{s\in \mathrm{val}(S)} -\ln\left[\beta_N^{p^\eta}(MI(p(\omega_N, A, B|s)))\right]$$

The first line is the BIC score. In the second line $\psi_2(N)$ is a weighting function such that $\psi_2(N)/N \to 0$ as $N \to \infty$: $\psi_2(N) := 1$ in the experiments. Each term in the sum is called a *sparsity boost*. The sum contains one sparsity boost for each *nonexistent* edge $(A, B) \notin G$. If $A \perp\!\!\!\perp B|(S = s)$, then the sparsity boost is $\Theta(N)$ as $N \to \infty$, and if $A \not\perp\!\!\!\perp B|(S = s)$, then it is $O(1)$, and further, in that case the sparsity boost becomes insignificant compared to the $LL$ term (since $\psi_2(N)/N \to 0$).

Second, the sets $S_{A,B}(G)$, called *separating sets*, are certain subsets of the power set of $V - \{A, B\}$, which provide *certificates for statistical recovery of* $G$. More precisely, we have $(A, B) \notin G$, if and only if there is a *witness* $S \in S_{A,B}(G)$ such that $A \perp\!\!\!\perp B|S$. The most common ways of defining $S_{A,B}(G)$ are as follows:

$$S_{A,B}(G) = \{S \subset V\backslash\{A, B\} \mid |S| \leq d\}, \tag{3}$$
$$S_{A,B}(G) = \{\mathrm{Pa}_G(A)\backslash B,\ \mathrm{Pa}_G(B)\backslash A\}. \tag{4}$$

The family of assignments $(A, B, G) \mapsto S_{A,B}(G)$ for all $(A, B)$ ranging over distinct pairs of vertices and $G$ over some family $\mathcal{G}$ of DAGs, constitutes a *collection of separating sets*, denoted by $\boldsymbol{\mathcal{S}}$.

In order for $A \perp\!\!\!\perp B|S$ to hold, we must have $A \perp\!\!\!\perp B|s$, for *every* joint assignment $s \in \mathrm{Val}(S)$. This is the reason for taking the minimum over $s \in \mathrm{Val}(S)$ of the possible sparsity boosts. The existence of just *one* $S \in S_{A,B}(G)$ such that $A \perp\!\!\!\perp B|S$ suffices to rule out $(A, B)$ as an edge in $G$. This is the reason for taking the maximum over $S \in S_{A,B}(G)$. The sparsity boost is $O(1)$ for an $(A, B) \in G$, and $\Theta(N)$ for an $(A, B) \notin G$.

It remains to explain how to compute $\beta_N^{p^\eta}(\gamma)$ in the implementation of the score $S_{\eta,\psi_1,\psi_2}$. According to the definition (2), $p^\eta$ is data-dependent, and this makes it impractical to compute $\beta_N^{p^\eta}(MI(\omega_N))$ quickly enough for use in our algorithm. We make an approximation by fixing $p^\eta$ to be a single "reference" distribution, with uniform marginals and satisfying $MI(p^\eta) = \eta$. In the case when $\mathrm{Val}(X_i) = \{0, 1\}$, there are two such distributions. Namely, let $p^0$ denote the uniform distribution, and let

$$p^0(t) = \begin{bmatrix} \frac{1}{4}+t & \frac{1}{4}-t \\ \frac{1}{4}-t & \frac{1}{4}+t \end{bmatrix} \text{ for all } t \in \left(-\frac{1}{4}, \frac{1}{4}\right). \quad (5)$$

Clearly, $p^0(t)$ has uniform marginals. Consider the function $MI(p^0(t))$ for $t \in \left(0, \frac{1}{4}\right)$. On this interval $MI(p^0(t))$ is positive, increasing, and has range $\left(0, MI\left(p^0\left(\frac{1}{4}\right)\right)\right)$. Thus for each $\eta$ in the range, there is a unique parameter value $t_\eta^+$ such that $MI(p^0(t_\eta^+)) = \eta$. By symmetry, we also have $MI(p^0(-t_\eta^+))) = \eta$; fix

$$p^\eta := p^0(t_\eta^+). \quad (6)$$

We compute $t_\eta^+$ by a binary search in the interval $\left(0, \frac{1}{4}\right)$; by (5) and (6) this suffices to compute $p^\eta$, and has to be done only once during the algorithm's setup.

Having computed $p^\eta$, we can compute $\beta_N^{p^\eta}(\gamma)$ for many values of $N, \gamma$, and store them in a table. During the learning phase, we evaluate $\beta_N^{p^\eta}(MI(\omega_N))$ by interpolation. We explain more details in Sec. 5.

**Related work**. Our new score is similar to other "hybrid" algorithms that use both conditional independence tests and score-based search for structure learning, notably Fast 2010's Greedy Relaxation algorithm (Relax) and Tsamardinos *et al.* 2006's Max-Min Hill-Climbing (MMHC) algorithm. The MMHC algorithm has two stages, first using independence tests to construct a skeleton of the Bayesian network, and then performing a greedy search over orientations of the edges using the BDeu score. The Relax algorithm starts by performing conditional independence tests to learn constraints, followed by edge orientation to produce an initial model. After the first model has been identified, Relax uses a local greedy search over possible relaxations of the constraints, at each step choosing the single constraint which, if relaxed, leads to the largest improvement in the score. Both of these algorithms separate the constraint- and score-based approaches into two distinct steps, in contrast to our approach which directly incorporates the conditional independence tests as a term in the score itself.

The only other work that we are aware of that has studied the incorporation of reliability of independence tests in score-based structure search is de Campos (2006). Their objective function is very different from ours, comparing the empirical mutual information to its expected value assuming independence (using the $\chi^2$ distribution). In contrast to de Campos's MIT score, the SparsityBoost score is consistent, provably able to recover the true structure.

**Importance of using Type II error.** To our knowledge, all previous approaches for Bayesian network structure learning use the Type I error $\alpha_N$ in assessing the reliability of an independence test, which is asymptotically given by the $\chi^2$ distribution. A relatively high threshold needs to be specified in order to prevent the false rejection of independence and to correct for multiple hypothesis testing. One of our key contributions is to show how to use $\beta_N$, the Type II error. *Minimizing the Type II error is essential because we want to err on the side of caution*, only having a large sparsity boost if we are sure that the corresponding edge does not exist. Type I errors, on the other hand, can be corrected by the part of the objective corresponding to the BIC score. If we had instead used the Type I error probability within our score, it would have corresponded to a dependence boost rather than independence, and would be fooled if we failed to find a good separating set (e.g., for computational reasons).

# 4 Polynomial Sample Complexity of the SparsityBoost Score

## 4.1 Statement of Main Results

In this section, we prove the consistency of the SparsityBoost score. In order to state our main results, we need to define certain additional parameters. First, there is a (small) positive integer, $d$, which bounds the in-degree of all vertices in $G$. The family of BNs on $n$ vertices satisfying this condition is called $\mathcal{G}^d$.

Second, we formalize the notion of the minimal edge strength $\epsilon$ in $G$. Define

$$S_{A,B}(\mathcal{G}^d) := \bigcup_{G \in \mathcal{G}^d} S_{A,B}(G).$$

Recall that the witness sets in $\mathcal{S}$ provide certificates for statistical recovery of $G$. We quantify the edge strength of $(A, B) \in G$ with respect to $S_{A,B}(\mathcal{G}^d)$, i.e.

the amount of dependence even after conditioning, by

$$\epsilon((A,B),S_{A,B}(\mathcal{G}^d)) := \min_{S\in S_{A,B}(\mathcal{G}^d)} \max_{s\in \mathrm{Val}(S)} MI(p(A,B|s))$$

Then, let $\epsilon = \epsilon(G) = \min_{(A,B)\in G} \epsilon((A,B),S_{A,B}(\mathcal{G}^d))$.

Next, we need the notion of an *error tolerance* $\zeta > 0$, which in turn follows from a notion of a $G' \in \mathcal{G}^d$ being $\zeta$-far from the true network $(G,P)$. For any $G' \in \mathcal{G}^d$, define the probability distribution $p_{G',P}$ over $\mathcal{X}$ to be the distribution which factors according to $G'$ and minimizes the KL-divergence from $P$, i.e.

$$p_{G',P} := \operatorname*{argmin}_{Q\,:\,G' \text{ is an I-map for } Q} H(P\|Q).$$

We call $H(P\|p_{G',P})$ the **divergence of $P$ from $G'$,** and if $H(P\|p_{G',P}) > \zeta$ we say that **$G'$ is $\zeta$-far from $(G,P)$**. In Theorem 1(a) we set an **error tolerance of $\zeta$,** which is to say that we specify that our learning algorithm should rule out all $G'$ which are $\zeta$-far from $(G,P)$.

Finally, we need $m$, the (maximum) inverse probability of an assignment to a separating set. More precisely, for any $A,B \in V^2$, $A \neq B$, and $S \in S_{A,B}(\mathcal{G}^d)$, let $m_P(S) := \max_{s\in\mathrm{Val}(S)}[P(S=s)]^{-1}$. Then let

$$m = m_P(G,\mathcal{G}^d,\mathcal{S}) = \max_{(A,B)\in G} \max_{S\in S_{A,B}(\mathcal{G}^d)} m_P(S). \quad (7)$$

For all $(A,B) \notin G$, there will be at least one witness $S \in S_{A,B}(G)$ such that $A \perp\!\!\!\perp B|S$. Let

$$\hat{S}_P((A,B),G) := \operatorname*{argmin}_{S\in S_{A,B}(G)\,:\,A\perp\!\!\!\perp B|S} m_P(S).$$

Finally, let

$$\hat{m}_P(G,\mathcal{S}) := \max_{(A,B)\notin G} m_P(\hat{S}_P((A,B),G)). \quad (8)$$

**Theorem 1** *Suppose that $(G,P) \in \mathcal{G}^d$ is a Bayesian network of $n$ **binary** random variables and $G$ is a perfect map for $P$. Set $S_{A,B}(G) = \{S \subset V\backslash\{A,B\} \mid |S| \leq d\}$. Assume that $(G,P) \in \mathcal{G}^d$ has minimal edge strength $\epsilon > 0$, and minimal assignment probabilities $m$, as defined in (7) and $\hat{m}_P(G,\mathcal{S})$, as defined in (8). Fix $\eta = \lambda\epsilon$ for $\lambda \in (0,1)$, an error probability $\delta > 0$, and a tolerance $\zeta > 0$. Let $S_\eta$ denote our score $S_{\eta,\psi_1,\psi_2}$ for $\psi_1(N) := \kappa\log(N)$ and $\psi_2(N) = 1$. Let $\omega_N$ be a sequence of observations sampled i.i.d. from $P$.*

*(a) There is a function $N(\epsilon,m,n;\delta,\zeta;\eta,\kappa)$ in*

$$\tilde{O}\left(\max\left(\frac{\log(n)m}{\epsilon^2},\ \frac{n^2}{\zeta^2}\right)\log\frac{1}{\delta}\right)$$

*as $\epsilon,\zeta,\delta \to 0^+$, $n,m \to \infty$, such that for all $N > N(\epsilon,m,n;\delta,\zeta;\eta,\kappa)$, with probability $1-\delta$, we have*

$$S_\eta(G,\omega_N) > S_\eta(G',\omega_N),$$

*for all $G' \in \mathcal{G}^d$ which are $\zeta$-far from $G$.*

*(b) Then there is a function $N(\epsilon,m,\hat{m}_P,n;\delta;\eta,\kappa)$ in*

$$\tilde{O}\left(\max\left(\frac{\log(n)m}{\epsilon^2},\ \frac{n^2\hat{m}_P^2}{\epsilon^2}\right)\log\frac{1}{\delta}\right)$$

*as $\epsilon,\delta \to 0^+$, $n,m,\hat{m}_P \to \infty$, such that for all $N > N(\epsilon,m,\hat{m}_P,n;\delta;\eta,\kappa)$, with probability $1-\delta$, we have*

$$S_\eta(G,\omega_N) > S_\eta(G',\omega_N),$$

*for all $G' \in \mathcal{G}^d$ such that* $\mathrm{Skel}(G') \not\subseteq \mathrm{Skel}(G)$.

In order to explain the significance of this result, it is helpful to relate it to three representative sample complexity results in the literature: Höffgen (1993), Friedman & Yakhini (1996), Zuk *et al.* (2006). The result of Zuk *et al.* differs from the other two and from our result because it only gives conditions for the (BIC) score of $G$ to beat that of an individual competing network $G'$, not a family, such as $\mathcal{G}^d$. The main difference between Höffgen and Friedman & Yakhini is that, like our result, Höffgen assumes that the competing network lies in $\mathcal{G}^d$ and achieves a sample complexity that is polynomial in $n = \mathrm{card}(V)$, while Friedman & Yakhini puts no restriction on the in-degree of competing networks, and obtains complexity that is exponential in $n$. Our result and Zuk *et al.* differ from both Höffgen and Friedman & Yakhini in that we provide guarantees for learning the correct DAG structure $G$ (or at least a $G$ without false edges), not just a distribution $P'$ which is $\zeta$-close to $P$. For this reason, only our paper and Zuk *et al.* need to define a minimal edge strength as a parameter, whereas for Höffgen and Friedman & Yakhini the main parameter is the error tolerance $\zeta$, which they call $\epsilon$.

### 4.2 Overview of Proofs

The proof of Theorem 1 consists of showing that for all sufficiently large $N$ we can find a (probable) lower bound on the *score difference*,

$$S_\eta(G,\omega_N) - S_\eta(G',\omega_N), \quad G,G' \in \mathcal{G}^d,\ \omega_N \sim G. \quad (9)$$

The score difference breaks down into a sum of the following terms:

(a) The difference of log-likelihood terms, $LL(G,\omega_N) - LL(G',\omega_N)$.

(b) The difference of complexity penalties, $\kappa\log(N)(|G'| - |G|)$.

(c) For each distinct pair of vertices $A,B \in V$ such that *neither* $G$ nor $G'$ has $(A,B)$ as an edge, the difference of the sparsity boosts in the objective functions of $G$ and $G'$, for that nonexistent edge.

(d) For each *true edge* $(A, B) \in G$ missing from $G'$, the negative of the sparsity boost for $(A, B) \notin G'$.

(e) For each *false edge* $(A, B) \notin G$ present in $G'$, the (positive) sparsity boost for $(A, B) \notin G$.

With the choice of $\mathcal{S}$ in the Theorem, $S_{A,B}(G') = S_{A,B}(G)$ for all $A, B \in V^2$, which implies that (c) is exactly 0. Furthermore, (b) is clearly $O(\log N)$ for $G, G' \in \mathcal{G}^d$, while both (a) and (e) will turn out to be $\Theta(N^\alpha)$ for $\alpha > 0$, so that (b) has only minor impact on the sample complexity.

So we will focus on how to estimate (a), (d), and (e). Conceptually, estimating each of these terms calls for the same *type of result*: a *concentration lemma* stating how quickly the empirical $LL(\cdot, \omega_N)$ (for (a)), respectively $MI(\omega_N, A, B|s)$ (for (d) and (e)) converges to the "ideal" counterpart $LL^{(I)}(\cdot, P)$, respectively $MI(P, A, B|s)$. In fact, both of the latter consist of a *polynomial* in $n$ number of terms (which is where we use the hypothesis $G \in \mathcal{G}^d$) of the form $p \log p$ for parameters $p$ of certain Bernoulli random variables.

**Proposition 1** *Let $p \in (0, 1)$ be given and $X(p)$ the Bernoulli random variable with parameter $p$. Let $\hat{\epsilon}, \delta \geq 0$ be given. For $Y_N \sim p$, denote the empirical parameter $p_{Y_N}$ by $\tilde{p}_N$. Then there is a function*

$$N(\hat{\epsilon}, \delta) \in O\left(\left(\frac{1}{\hat{\epsilon}}\right)^2 \log \frac{1}{\delta}\right), \tag{10}$$

*as $\hat{\epsilon}, \delta \to 0^+$ with the property that for $N > N(\hat{\epsilon}, \delta)$,*

$$\Pr\left(|\tilde{p}_N \log \tilde{p}_N - p \log p| < \hat{\epsilon}\right) \geq 1 - \delta.$$

Proposition 1 improves slightly on Lemma 1 in Höffgen (1993), by replacing $\tilde{O}(\cdot)$ with $O(\cdot)$ in (10).

A key feature of Proposition 1, for obtaining our concentration results for $LL$ and $MI$ is that (10) is independent of the Bernoulli parameter $p$. From the concentration result for $LL$, we can show that (a) is with high probability positive and larger than $N\zeta/3$, for all $G'$ which are $\zeta$-far from $G$ and for sufficiently large $N$. From the concentration result for $MI$ we can show that a sparsity boost from a *true* edge is bounded above by a constant for sufficiently large $N$ (linear in $m$). So the negative contribution of (d) is bounded. These bounds suffice to prove Theorem 1(a).

In the proof of Theorem 1(b), from the concentration result for $LL$, we can show that (a) is with high probability larger than a constant times $-n\sqrt{\log(n)N}$. Furthermore, a sparsity boost from a *false* edge is $\Omega(\Gamma(\eta)N)$, where the speed $\Gamma(\eta)$ of the linear growth is on the order of $\eta^2$ as $\eta \to 0^+$. To show the latter, we first apply Proposition 1, given a witness, to prove that $MI(\omega_N, A, B|s)$ is (likely) less than $\eta/2$. Second, using a Chernoff bound, we show that $-\log \beta_N^{p^\eta}(\gamma)$ is $\Omega(\eta^2 N)$ for $\gamma$ less than $\eta/2$. So, with high probability the positive contribution of (e) eventually overwhelms any negative contribution of (a).

The techniques derived from the Chernoff bound yield a version of Theorem 1(b) with an exponent of 4 on the $\epsilon$ in the denominator of the term $n^2 \hat{m}_P^2/\epsilon^2$. To improve the exponent to 2, we need a strengthened result on the linear growth of a sparsity boost from a false edge, in which the speed $\Gamma(\eta)$ is on the order of only $\eta$ instead of $\eta^2$, as $\eta \to 0^+$.

We have to use a new method derived from Sanov's Theorem instead of Chernoff's Bound. To our knowledge, the way we use Sanov's Theorem to study the concentration of mutual information is a novel contribution to information theory. For all of the following we are assuming that $\mathrm{Val}(X_i) = \{0, 1\}$ for all $X_i \in V$ so that $\mathcal{P}$ is the space of probability distributions over the alphabet $\{0, 1\}^2$. We have already, in (5), given a parameterization of the path of distributions of uniform marginals in $\mathcal{P}$. We now generalize (5) and the associated parameterization by defining

$$p(p_{A,0}, p_{B,0}, t) := \begin{bmatrix} p_{A,0}p_{B,0} + t & p_{A,1}p_{B,0} - t \\ p_{A,0}p_{B,1} - t & p_{A,1}p_{B,1} + t \end{bmatrix} \tag{11}$$

where $p_{A,1} := 1 - p_{A,0}$ and $p_{B,1} := 1 - p_{B,0}$. When $(p_{A,0}, p_{B,0})$ ranges over $[0, 1]^2$ and $t$ over $(t_{\min}, t_{\max})$ (an interval depending on $p_{A,0}$, $p_{B,0}$), (11) parameterizes the whole space $\mathcal{P}$.

Since the $t$ parameter is a measure of how far $p$ is from $\mathcal{P}_0$, it is not surprising that we can derive formulas relating $t$ to $\sqrt{MI}$. In order to carry this out, we consider the function $MI(p(p_{A,0}, p_{B,0}, t))$ as a function of $t$ and carefully study the Taylor series expansion of this function around the basepoint $t = 0$.

The reason for preferring the $t$ parameter to $MI$ itself is that by means of Sanov's Theorem and Pinsker's Inequality, we obtain a very general result which bounds $-\log \beta_N^{p^\eta}(\gamma)$ from below by $N$ times the squared $L_\infty$-distance of $p^\eta$ from a distribution $q^\gamma$. More specifically, defining the complement of $P_\gamma$ by

$$A_\gamma := \{p \in \mathcal{P} \mid MI(p) \leq \gamma\}, \tag{12}$$

the distribution $q^\gamma$ is defined as the I-projection of $p^\eta$ onto $A_\gamma$. We would like to relate $\|p^\eta - q^\gamma\|_\infty$ to $|MI(p^\eta) - MI(q^\gamma)| = |\eta - \gamma|$, and the $t$-parameters act as an effective intermediary, because it is easy to show that $\|p^\eta - q^\gamma\|_\infty$ is on the order of $|t_\eta^+ - t_\gamma^+|$, where $t_\gamma^+$ is the $t$-parameter of $q^\gamma$. Applying the relation of the preceding paragraph between $t$ and $\sqrt{MI}$, we obtain a bound, from below, of $-\log \beta_N^{p^\eta}(\gamma)$ by something on the order of $(\sqrt{\eta} - \sqrt{\gamma})^2 N$.

## 5 Computation of $\beta$ values

**Exact computation.** Here we give an exact formula for $\beta_N^{p^\eta}(\gamma)$ using the Method of Types (Cover & Thomas, 2006, Chapter 11). Denoting the entries of $p^\eta \in \mathcal{P}$ by $(p_{0,0}, p_{0,1}, p_{1,0}, p_{1,1})$, we have

$$\beta_N^{p^\eta}(\gamma) = \sum_{Y_N} \prod_{i,j=0}^{1} p_{i,j}^{T_{i,j}(Y_N)} \ \mathbf{1}[MI(Y_N) \leq \gamma],$$

where $T_{i,j}(Y_N)$ is the number of observations of $(i,j)$ in the sampled sequence $Y_N$ of length $N$. Consider the set $\mathcal{T}_N$ of length-4 vectors of nonnegative integers $(T_{0,0}, T_{0,1}, T_{1,0}, T_{1,1})$ summing to $N$. Every $T \in \mathcal{T}_N$ corresponds one-to-one with a distribution $p_T \in \mathcal{P}$ (obtained by dividing every entry in $T$ by $N$). Let $|T|$ denote the number of sequences $Y_N$ corresponding to type $T$. Then it is not difficult to see that $|T|$ is given by a multinomial coefficient and that

$$\beta_N^{p^\eta}(\gamma) = \sum_{T \in \mathcal{T}_N} |T| \prod_{i,j=0}^{1} p_{i,j}^{T_{i,j}} \mathbf{1}_{A_\gamma}(p_T), \qquad (13)$$

where $\mathbf{1}_{A_\gamma}$ is the characteristic function of $A_\gamma$ (see Eq. 12). We can use (13) to exactly compute $\beta_N^{p^\eta}(\gamma)$, but because of the summation over $\mathcal{T}_N$ the running time of this algorithm is $O(N^3)$, which will not scale to the range of $N$ we need for our experiments.

**Monte Carlo computation.** In place of exact calculation, we estimate $\beta_N^{p^\eta}$ by means of Monte Carlo integration, using importance sampling of the domain to reduce the variance. In order to implement this, we first observe that (13) is essentially a Riemann sum for a definite integral, so that we may replace the summation with an integral. Second, the integrand we initially obtain in this manner has numerous discontinuities, because of the $|T|$ factor. It makes the next steps easier to implement if we replace $|T|$ with a (slightly larger) continuous approximation (Csiszar & Körner, 2011, p. 39). We finally obtain the following integral which approximates $\beta_N^{p^\eta}(\gamma)$ given in (13):

$$\left(\frac{N}{2\pi}\right)^{(|X|-1)/2} \int_{\mathcal{P}} e^{-NH(q\|p^\eta)} \prod_{i,j=0}^{1} q_{i,j}^{-\frac{1}{2}} \ \mathbf{1}_{A_\gamma}(q)\, \mathrm{d}q.$$

For the Monte Carlo integration we use an importance sampling scheme based on the following idea: the integrand is largest when $H(q\|p^\eta)$ is small and $q \in A_\gamma$, and so it should be strongly concentrated around $q^\gamma$ (the I-projection of $p^\eta$ onto $A_\gamma$). We have an unproven conjecture, supported by numerical evidence, that $q^\gamma = p^\gamma := p^0(t_\gamma^+)$ (the unproven part of this is that $q^\gamma$ has uniform marginals) for $\eta$ less than approximately 0.1109. The importance sampling algorithm samples points $p \in \mathcal{P}$ i.i.d., favoring points near $p^\gamma$. We use the parameterization (11) of $\mathcal{P}$ and sample the parameters $p_{A,0}, p_{B,0}$ & $t$ independently according to Gaussian distributions. For the selection of the two marginals, we use identical Gaussians centered at $\frac{1}{2}$ and becoming more concentrated (exponentially fast) around their mean as $N \to \infty$. For the $t$ parameter, we use use a third Gaussian centered at $t_\gamma^+$. For each $(N,\gamma)$ we determine the concentration of the third Gaussian by sampling the integrand along the path $p\left(\frac{1}{2}, \frac{1}{2}, t\right)$, in the segment $(0, t_\gamma^+)$.

Since we cannot possibly tabulate $\beta_N^{p^\eta}(\omega_N)$ for every empirical sequence that might arise, we tabulate $\beta_N^{p^\eta}(\gamma)$ for $N, \gamma$ in a strategically chosen grid of values, and during the learning phase we interpolate or extrapolate (as the need arises) from these tabulated values. We interpolate/extrapolate $-\ln \beta_N^{p^\eta}(\gamma)$ *linearly* in the statistics $N$ and $H(p^\gamma\|p^\eta)$. Sanov's Theorem gives heuristic support to this procedure, but ultimately our justification for this procedure rests on the empirical results presented in Section 6 below.

## 6 Experimental Results

**Computing the confidence measure.** In Figure 1 we present several empirical results that help to justify our methods for calculating $\beta_N^{p^\eta}(\gamma)$, our new measure of the reliability of an independence test. First, in (a), we show that using the method of summing over types to calculate $\beta_N^{p^\eta}(\gamma)$ has a running time which is $O(N^4)$, whereas the Monte Carlo method explained in Section 5 is $O(1)$ as $N \to \infty$. Thus, although it is feasible to pre-compute $\beta_N^{p^\eta}(\gamma)$ for small values of $N$, exact calculation is impractical for $N$ much larger than 200.

As for the accuracy of the Monte Carlo estimation, the table in Figure 2 shows that for very small $N$, e.g. $N < 50$, some multiplicative errors for our method of $\approx 30\%$ are observed, but by the time we reach $N = 100$, the errors are consistently $< 10\%$. Figure 1(b) shows that, for $N = 200$, the Monte Carlo estimate has a consistently small error over the range of $\gamma$.

The linear interpolation procedure for obtaining $\ln \beta_N^{p^\eta}(MI(\omega_N))$ from the pre-computed tables of $\ln \beta_N^{p^\eta}(\gamma)$ receives heuristic support from Sanov's Theorem; it receives empirical support from Figure 1(c) (resp. (d)), which shows that the dependence of $\ln \beta_N^{p^\eta}(\gamma)$ on $N$ (resp., $H(p^\gamma\|p^\eta)$), assuming all other inputs are fixed, is roughly linear.

**Sample Complexity.** In this section we study the accuracy of our learning algorithm as a function of the amount of data we provide it. We compare our algorithm to two baselines: BIC and Max-Min Hill-Climbing. BIC is equivalent to our score without the sparsity boost terms. MMHC is state-of-the-art in terms of both speed and quality of recovery, and has been shown to outperform most other constraint-based approaches (Tsamardinos *et al.* , 2006). As we discussed earlier, MMHC is also a hybrid algorithm, using both conditional independence tests and score-based search. We use the implementation of MMHC provided by the authors as part of Causal Explorer 1.4 (Aliferis *et al.* , 2003), with the default parameters.[1]

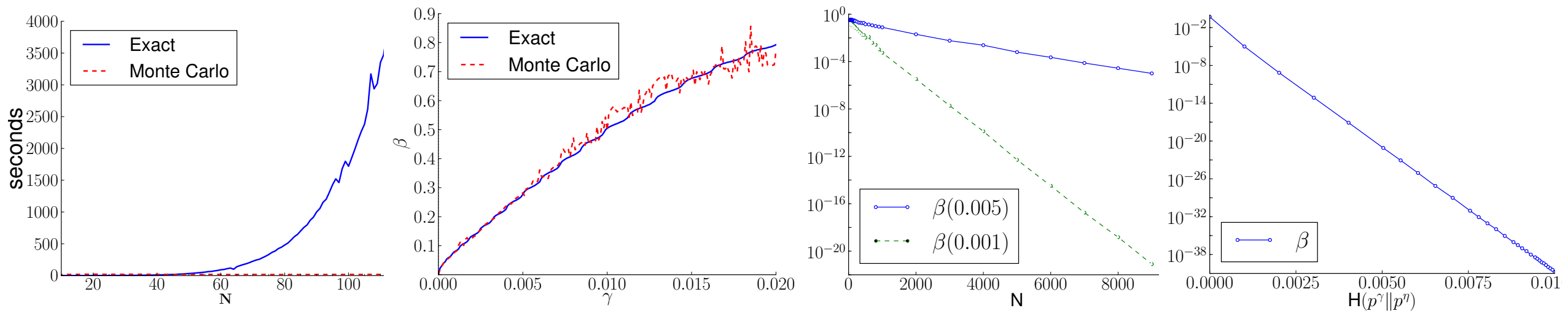

(a) (b) (c) (d)

Figure 1: Computation of $\beta_N^{p^\eta}$. All results shown are for $\eta = 0.01$. (a) Running time of the exact algorithm to compute $\beta_N^{p^\eta}$ grows cubically in $N$, but for Monte Carlo approximation remains constant (results shown for $\gamma = 0.005$ and 0.001 combined). (b) Monte Carlo estimate of $\beta_N^{p^\eta}(\gamma)$ for fixed $\eta$, $N = 200$. (c) Exponential decay of $\beta_N^{p^\eta}(\gamma)$ in $N$ for fixed $\gamma$. (d) Exponential decay of $\beta_N^{p^\eta}(\gamma)$ as a function of KL-divergence $H(p^\gamma \| p^\eta)$, as $\gamma$ is varied, for large $N = 9000$.

| | $\gamma$ | | | $\gamma$ | |
|---|---|---|---|---|---|
| N | 0.001 | 0.005 | N | 0.001 | 0.005 |
| 20 | 12.20% | 29.15% | 70 | 1.03% | 3.59% |
| 30 | 24.65% | 1.18% | 80 | 9.06% | 4.13% |
| 40 | 39.77% | 7.07% | 90 | 0.77% | 4.05% |
| 50 | 2.45% | 4.88% | 100 | 1.01% | 0.03% |
| 60 | 3.52% | 0.30% | 110 | 2.27% | 3.01% |

Figure 2: Multiplicative error of Monte Carlo approximation, $|\beta_N^{p^\eta} - \hat{\beta}_N^{p^\eta}|/\beta_N^{p^\eta}$, for $\eta = 0.01$ as $N, \gamma$ vary.

We use an integer linear program to exactly solve for the Bayesian network that maximizes the BIC or SparsityBoost scores (Jaakkola *et al.* , 2010; Cussens, 2011). To solve the ILP, we use Cussens' GOBNILP 1.2 software together with SCIP 3.0 (Achterberg, 2009). Conveniently, since the sparsity boost terms in our objective can be subsumed into the parent set scores, we can use these off-the-shelf Bayesian network solvers without any modification.

The data that we use for learning is sampled from synthetic distributions based on the Alarm network structure (Beinlich *et al.* , 1989). The Alarm network has 37 variables, 46 edges, and a maximum in-degree of 4. In our synthetic distributions, every variable has only two states, and its conditional probability distribution is given by a logistic function, $p(X_i = 1 \mid \mathbf{x}_{\text{Pa}(i)}) = 1/(1 + e^{-\vec{\theta}_i \cdot \mathbf{x}_{\text{Pa}(i)} - u_i})$. We sampled 10 different distributions, with parameters drawn according to $\theta_{ij} \sim U[-.5, .5] + \frac{1}{4}\mathcal{N}(0, 1)$ for $j \in \text{Pa}(i)$ and $u_i \sim \frac{1}{4}\mathcal{N}(0, 1)$. For each value of $N$, a new set of $N$ samples were drawn from the corresponding synthetic distribution. The results shown are the average for each of these 10 synthetic distributions.

[1] Threshold for $\chi^2$ test of .05 and Dirichlet hyperparameters equal to 10. Varying these did not improve results.

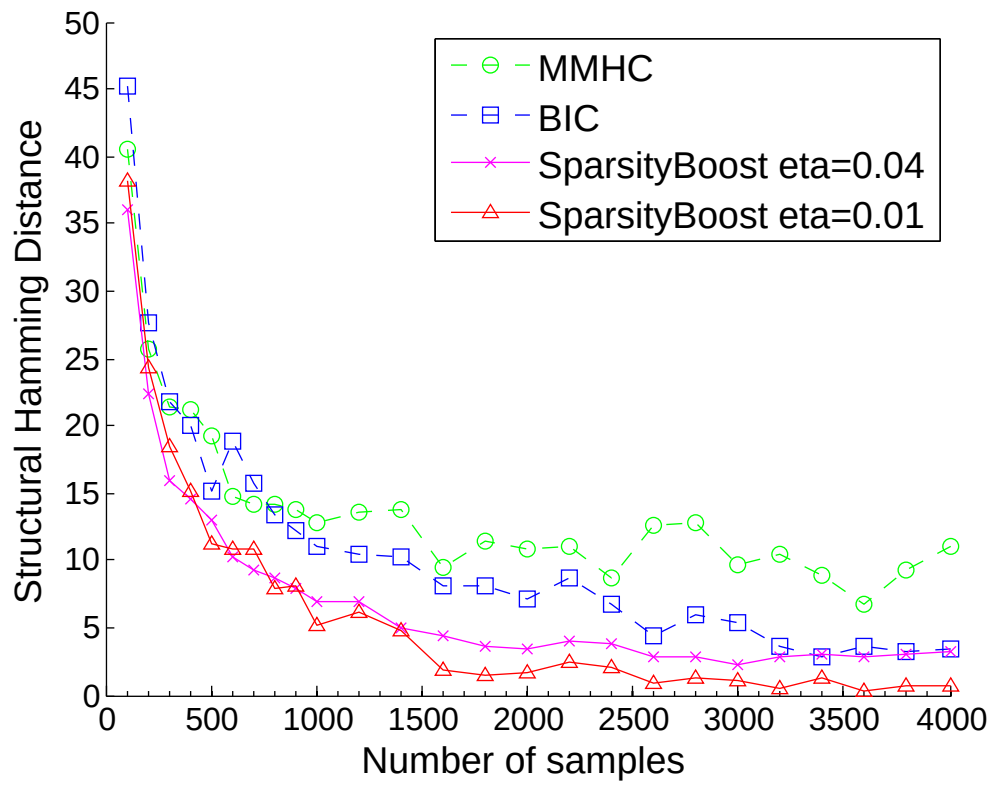

Figure 3: Comparison of the sample complexity of MMHC, BIC, and our new SparsityBoost objective. Each point is the average of the SHD of the learned network from truth for 10 synthetic distributions.

We use $S_{A,B}(G)$ from Eq. 3 with $d = 2$, enumerating over all separating sets of size at most two. Larger separating sets are less useful because they lead to a smaller $\epsilon$, less data, and more computation to create the objective. In the Alarm network, for every $(A, B) \notin G$ there is a separating set $S$ such that $|S| \leq 2$ and $A \perp B | S$. Regardless, if a separating set for an inexistent edge cannot be found, our score simply reduces to the BIC score, so no harm is done.

Our results are shown in Figure 3. We measure the quality of structure recovery using the Structural Hamming Distance (SHD) between the partially directed acyclic graphs (PDAG) representing the equivalence classes of the true and learned networks (Tsamardinos *et al.* , 2006). The SHD is defined as the number of edge additions, deletions, or reversals to make the two PDAGs match. The plots for SparsityBoost with $\eta = 0.005$ and $\eta = 0.02$ (not shown) are nearly identical to that of $\eta = 0.01$. SparsityBoost consistently learns better structures than MMHC or BIC, and often perfectly recovers the networks after only 1600 samples. SparsityBoost obtains a smaller average error with 3000 samples than BIC does with 6000, representing a more than 50% reduction in the number of samples needed for learning. We also found that the SparsityBoost results had substantially less variance than either BIC or MMHC.

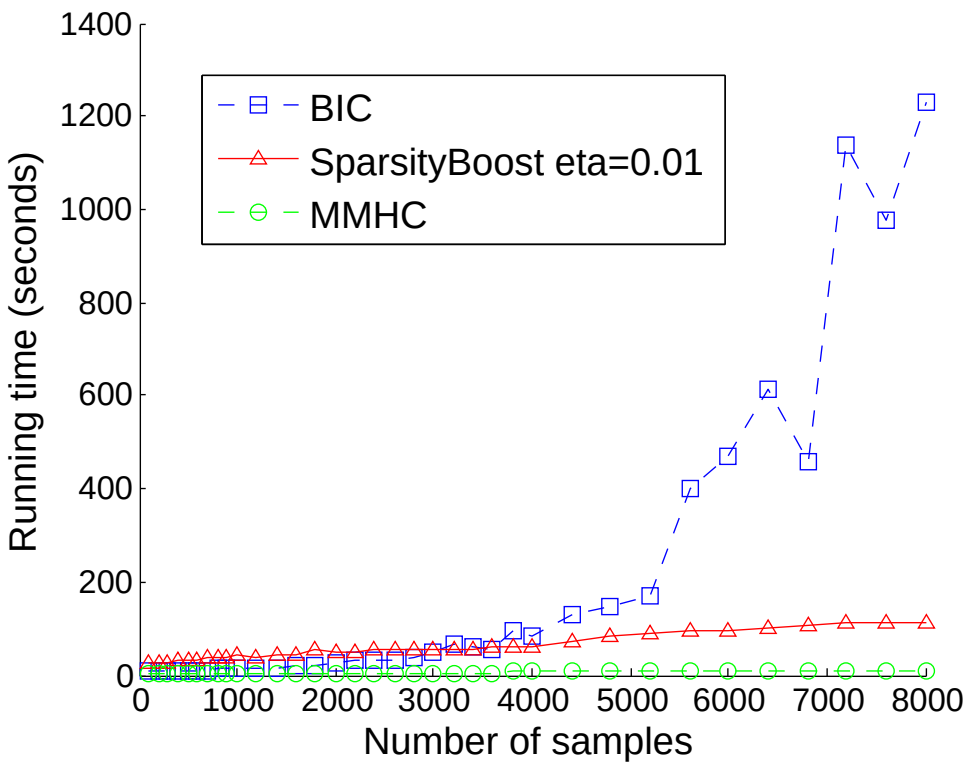

Figure 4: Total running time to learn a Bayesian network from data for BIC, SparsityBoost, and MMHC. We maximize the BIC and SparsityBoost scores by solving an integer linear program to optimality.

Our theoretical results only guarantee exact recovery when $\eta < \epsilon$. For each of the synthetic distributions we computed $\epsilon(A, B)$ for all of the edges $(A, B)$ in the true structure (see Sec. 4.1 for definition). The minimum of these, that is to say $\epsilon$, ranged from .000028 to .0047, which is in fact *smaller* than the largest value of $\eta$ considered in our experiments (.005). Despite this, we obtained excellent empirical results for SparsityBoost with $\eta \in \{.005, .01, .02\}$. This may be partially explained by the *average* value of $\epsilon(A, B)$ being .062. Even when we push $\eta$ to be as high as .04, SparsityBoost converges to an average SHD of at most 3 (see Fig. 3). Thus, our new objective appears to be particularly robust to choosing the wrong value of $\eta$.

**Running Time.** We show the running time of our new objective compared to BIC in Figure 4. The figure shows the total time, which includes both the time to compute the score of all parent sets and the time to solve the ILP to optimality. These results confirm our hypothesis that the new score would be substantially easier to optimize. We found that the linear programming relaxation for SparsityBoost (with $\eta = 0.01$) was tight on *nearly all* instances: branch-and-bound did not need to be performed. Once the SparsityBoost objective has been computed, the ILP is solved within 6 seconds in every single instance.

The timing experiments reported in this section were performed on a single core of a 2.66 Ghz Intel Core i7 processor with 4 GB of memory. MMHC's average running time was less than 8 seconds for all sample sizes. MMHC is significantly faster because it quickly prunes edges that cannot exist and in its second step uses a greedy (rather than exact) optimization algorithm for score-based search.

## 7 Discussion

Our approach maintains the advantages of other score-based approaches to structure learning, such as the ability to find the $K$-best Bayesian networks and ease of introducing additional constraints (e.g., from interventional data). In order to optimize our score, virtually any optimization procedure can be used. Since the ILP is easy to solve, this suggests that greedy structure search may also be able to easily find the best-scoring Bayesian network under the SparsityBoost score.

One subject for future investigations is to generalize and sharpen our results in various ways. Using a similar construction for $p^\eta$, we believe it should be possible to extend our score and proof of consistency to non-binary variables. We also believe it will be possible to eliminate the dependence of $N(\epsilon, m, \hat{m}_P, n; \delta, \zeta; \eta, \kappa)$ in both parts of Theorem 1 on the parameter $m$, leaving only the dependence on $\hat{m}_P$ in part (b), which is in some cases much smaller than $m$.

Another issue to be explored as a future line of investigation is the choice of $p^\eta$ in our measure of reliability, $\beta_N^{p^\eta}(\gamma)$. The overall motivation for $\beta_N^{p^\eta}(\gamma)$ is to capture the probability of Type II error of a statistical test with independent distributions $\mathcal{P}_0$ as the null hypothesis $H_0$ and all distributions $\mathcal{P}_\eta$ as the alternative hypothesis $H_1$. The choice of uniform marginals for $p^\eta$ represents an expedient choice, providing an objective function that is manageable to implement and compute, yet still has a reasonable theoretical and empirical sample complexity. Better results might be obtained by setting the marginals of $p^\eta$ to approximate those of $p(\omega_N)$. More generally, one can contemplate incorporating various other statistically derived probabilities into the objective function.

This leads to the broader point that objective functions, and the optimization of them over discrete spaces of structures, are ubiquitous throughout computer science and statistics. Our work suggests a new paradigm for incorporating information from "classical" hypothesis tests into the objective functions used for machine learning. This new paradigm provides both computational *and* statistical efficiency.

## References

Achterberg, Tobias. 2009. SCIP: Solving constraint integer programs. *Mathematical Programming Computation*, **1**(1), 1–41.

Aliferis, Constantin F., Tsamardinos, Ioannis, Statnikov, Alexander R., & Brown, Laura E. 2003. Causal Explorer: A Causal Probabilistic Network Learning Toolkit for Biomedical Discovery. *Pages 371–376 of: METMBS.*

Beinlich, I. A., Suermondt, H. J., Chavez, R. M., & Cooper, G. F. 1989. The ALARM Monitoring System: A Case Study with Two Probabilistic Inference Techniques for Belief Networks. *Pages 247–256 of: Proceedings of the 2nd European Conference on Artificial Intelligence in Medicine*. Springer-Verlag.

Chickering, D. 1996. Learning Bayesian Networks is NP-Complete. *Pages 121–130 of:* Fisher, D., & Lenz, H.J. (eds), *Learning from Data: Artificial Intelligence and Statistics V*. Springer-Verlag.

Chickering, D. 2002. Learning Equivalence Classes of Bayesian-Network Structures. *Journal of Machine Learning Research*, **2**, 445–498.

Chickering, D., Heckerman, D., & Meek, C. 2004. Large-Sample Learning of Bayesian Networks is NP-Hard. *Journal of Machine Learning Research*, **5**, 1287–1330.

Cover, T.M., & Thomas, J.A. 2006. *Elements of information theory.* John Wiley & Sons.

Csiszar, I., & Körner, J. 2011. *Information theory: coding theorems for discrete memoryless systems.* Cambridge University Press.

Cussens, James. 2011. Bayesian network learning with cutting planes. *Pages 153–160 of: Proceedings of the Twenty-Seventh Conference Annual Conference on Uncertainty in Artificial Intelligence (UAI-11).* Corvallis, Oregon: AUAI Press.

Dasgupta, S. 1999. Learning polytrees. *In: Proc. of the 15th Conference on Uncertainty in Artificial Intelligence.*

de Campos, Luis M. 2006. A Scoring Function for Learning Bayesian Networks based on Mutual Information and Conditional Independence Tests. *J. Mach. Learn. Res.*, **7**(Dec.), 2149–2187.

Dembo, A., & Zeitouni, O. 2009. *Large deviations techniques and applications.* Vol. 38. Springer.

Fast, A. 2010. *Learning the structure of Bayesian networks with constraint satisfaction.* Ph.D. thesis, University of Massachusetts Amherst.

Friedman, N., & Yakhini, Z. 1996. On the Sample Complexity of Learning Bayesian Networks. *In: Proc. of the 12th Conference on Uncertainty in Artificial Intelligence.*

Friedman, Nir, Nachman, Iftach, & Pe'er, Dana. 1999. Learning Bayesian Network Structure from Massive Datasets: The "Sparse Candidate" Algorithm. *Pages 206–215 of: UAI.*

Heckerman, D., Geiger, D., & Chickering, D. M. 1995. Learning Bayesian Networks: The Combination of Knowledge and Statistical Data. *Machine Learning*, **20**(3), 197–243. Available as Technical Report MSR-TR-94-09.

Hoeffding, W. 1965. Asymptotically optimal tests for multinomial distributions. *The Annals of Mathematical Statistics*, 369–401.

Höffgen, K.U. 1993. Learning and robust learning of product distributions. *Pages 77–83 of: Proceedings of the sixth annual conference on Computational learning theory.* ACM.

Jaakkola, Tommi, Sontag, David, Globerson, Amir, & Meila, Marina. 2010. Learning Bayesian Network Structure using LP Relaxations. *Pages 358–365 of: Proceedings of the Thirteenth International Conference on Artificial Intelligence and Statistics (AI-STATS)*, vol. 9. JMLR: W&CP.

Lam, Wai, & Bacchus, Fahiem. 1994. Learning Bayesian Belief Networks: An Approach Based on the MDL Principle. *Computational Intelligence*, **10**, 269–294.

Paninski, Liam. 2003. Estimation of entropy and mutual information. *Neural Comput.*, **15**(6), 1191–1253.

Pearl, Judea, & Verma, Thomas. 1991. A Theory of Inferred Causation. *Pages 441–452 of: KR.*

Sachs, Karen, Perez, Omar, Pe'er, Dana, Lauffenburger, Douglas A., & Nolan, Garry P. 2005. Causal Protein-Signaling Networks Derived from Multiparameter Single-Cell Data. *Science*, **308**(5721), 523–529.

Spirtes, P., Glymour, C., & Scheines, R. 2001. *Causation, Prediction, and Search, 2nd Edition.* The MIT Press.

Tsamardinos, I., Brown, L. E., & Aliferis, C. F. 2006. The Max-Min Hill-Climbing Bayesian Network Structure Learning Algorithm. *Machine Learning*, **65**(1), 31–78.

Zuk, Or, Margel, Shiri, & Domany, Eytan. 2006. On the Number of Samples Needed to Learn the Correct Structure of a Bayesian Network. *In: UAI.* AUAI Press.